\documentclass[10pt]{article}

\usepackage{lineno,hyperref}
\modulolinenumbers[5]

\usepackage[margin=0.75in]{geometry}
\usepackage{amsmath,amsthm, amssymb}
\usepackage[all]{xy}
\usepackage{graphicx}

\newtheorem{myconjecture}{Conjecture}

\begin{document}


\title{Toward a Formal Model of Cognitive Synergy}

\author{Ben Goertzel \\
OpenCog Foundation, Hong Kong \\
ben@goertzel.org}




\maketitle

\begin{abstract}
"Cognitive synergy" refers to a dynamic in which multiple cognitive processes, cooperating to control the same cognitive system, assist each other in overcoming bottlenecks encountered during their internal processing.   Cognitive synergy has been posited as a key feature of real-world general intelligence, and has been used explicitly in the design of the OpenCog cognitive architecture.  Here category theory and related concepts are used to give a formalization of the cognitive synergy concept.  

A series of formal models of intelligent agents is proposed, with increasing specificity and complexity: simple reinforcement learning agents; "cognit" agents with an abstract memory and processing model; hypergraph-based agents (in which "cognit" operations are carried out via hypergraphs); hypergraph agents with a rich language of nodes and hyperlinks (such as the OpenCog framework provides); "PGMC" agents whose rich hypergraphs are endowed with cognitive processes guided via Probabilistic Growth and Mining of Combinations; and finally variations of the PrimeAGI design, which is currently being built on top of OpenCog.   

A notion of cognitive synergy is developed for cognitive processes acting within PGMC agents, based on developing a formal notion of "stuckness," and defining synergy as a relationship between cognitive processes in which they can help each other out when they get stuck.   It is proposed that cognitive processes relating to each other synergetically, associate in a certain way with functors that map into each other via natural transformations.  Cognitive synergy is proposed to correspond to a certain inequality regarding the relative costs of different paths through certain commutation diagrams.

Applications of this notion of cognitive synergy to particular cognitive phenomena, and specific cognitive processes in the PrimeAGI design, are discussed.
\end{abstract}



\tableofcontents

\section{Introduction}

General intelligence is a broad concept, going beyond the "g-factor" used to measure general intelligence in humans and broadly beyond the scope of "humanlike intelligence."   Whichever of the available formalizations of the "general intelligence" concept one uses \cite{LeggHutter2007, Legg2007a, Goertzel2010c},
leads to the conclusion that humanlike minds form only a small percentage of the space of all possible
generally intelligent systems.  This gives rise to many deep questions, including the one that motivates the present paper: {\bf Do there exist general principles, which {\it any} system must obey in order to achieve advanced general intelligence using feasible computational resources?}  

Psychology and neuroscience are nearly
mute on this point, since they focus on human and animal intelligence
almost exclusively.  The current mathematical theory of general intelligence
doesn't help much either, as it focuses mainly on the properties
of general intelligences that use massive, infeasible amounts of computational
resources \cite{Hutter2005}.  On the other hand, current practical AGI work focuses on specific classes of systems
that are hoped to display powerful general intelligence, and the level of genericity
of the underlying design principles is rarely clarified.  For instance, Stan Franklin's
AGI designs \cite{Baars2009} are based on Bernard Baars' Global Workspace theory \cite{Baars97}, 
which is normally presented as a model of human intelligence; it's unclear whether either
Franklin or Baars considers the Global Workspace theory to possess a validity beyond the 
scope of humanlike general intelligence.

So, this seemingly basic question about general principles of general intelligence
pushes beyond the scope of current AGI theory and practice, cognitive science
and mathematics.  This paper seeks to take a small step in the direction of an answer.

In \cite{EGI1} one possible general principle of computationally feasible general intelligence
was proposed -- the principle of "cognitive synergy."   The basic concept of cognitive synergy, as presented there, is that general intelligences must contain
different knowledge creation mechanisms corresponding to
different sorts of memory (declarative, procedural,
sensory/episodic, attentional, intentional); and that these
different mechanisms must be interconnected in such a
way as to aid each other in overcoming memory-type-specific
combinatorial explosions.

In this paper, cognitive synergy is revisited and given a more formal description in the language of category theory.
This formalization is a presented both for the conceptual clarification it offers, and as a hopeful step toward proving interesting theorems about the relationship between
cognitive synergy and general intelligence, and evaluating the degree of cognitive
synergy enabled by existing or future concrete AGI designs.  The relation of the formal notion of cognitive synergy presented to the OpenCog / PrimeAGI design developed by the author and colleagues \cite{EGI1} \cite{EGI2} is discussed in moderate detail, but this is only one among many possible examples; the general ideas proposed here should be applicable to a broad variety of AGI designs.

\section{Cognit Agents: A General Formalization of Intelligent Systems}

\begin{figure*}[htb]
\centering
\includegraphics[width=8cm]{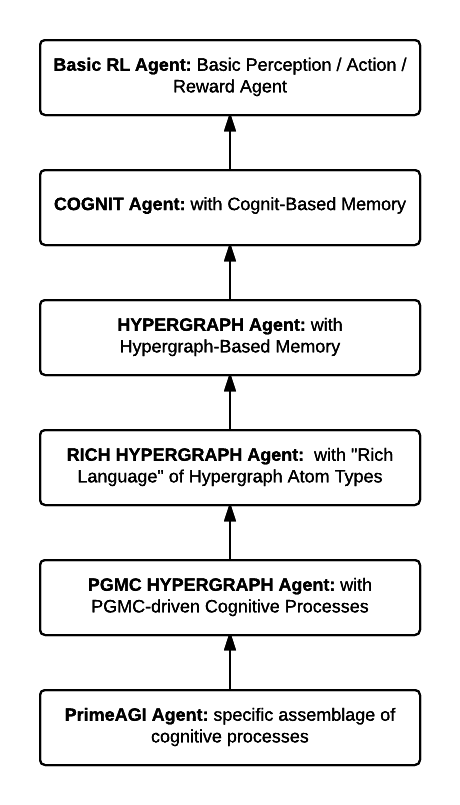}
\caption{An inheritance hierarchy showing the formal models of intelligent agents discussed here, with the most generic at the top and the most specific at the bottom.}
\label{fig:agent-classes}
\end{figure*}

We will introduce here a hierarchy of formal models of intelligent agents, beginning with a very simple agent that has no structure apart from the requirement to issue actions and receive perceptions and rewards; and culminating with a specific AGI architecture, PrimeAGI \footnote{The architecture now labeled PrimeAGI was previously known as CogPrime, and is being implemented atop the OpenCog platform} \cite{EGI1}\cite{EGI2}.  The steps along the path from the initial simple formal model toward OpenCog will each add more structure and specificity, restricting scope and making finer-grained analysis possible.   Figure \ref{fig:agent-classes} illustrates the hierarchy to be explored.

For the first step in our agent-model hierarchy, which we call a {\bf Basic RL Agent} (RL for Reinforcement Learning), we will follow \cite{Hutter2005, legg2008machine} and consider a model involving a class of active agents which observe and explore their environment and also take actions in it, which may affect the environment.  Formally, the agent in our model sends information to the environment by sending symbols from some finite alphabet called the {\it action space} $\Sigma$; and the environment sends signals to the agent with symbols from an alphabet called the {\it perception space}, denoted $\mathcal P$.  Agents can also experience rewards, which lie in the {\it reward space}, denoted $\mathcal R$, which for each agent is a subset of the rational unit interval.  

The agent and environment are understood to take turns sending signals back and forth, yielding a history of actions, observations and rewards, which may be denoted

$$
a_1 o_1 r_1 a_2 o_2 r_2 ...
$$

\noindent or else $a_1 x_1 a_2 x_2 ...$ if $x$ is introduced as a single symbol to denote both an observation and a reward.  The complete interaction history up to and including cycle $t$ is denoted $ax_{1:t}$; and the history before cycle t is denoted $ax_{<t}$ = $ax_{1:t-1}$.

The agent is represented as a function $\pi$ which takes the current history as input, and produces an action as output.  Agents need not be deterministic, an agent may for instance induce a probability distribution over the space of possible actions, conditioned on the current history.  In this case we may characterize the agent by a probability distribution $\pi( a_t  | ax_{<t} )$.    Similarly, the environment  may be characterized by a probability distribution $\mu(x_k | ax_{<k} a_k)$.  Taken together, the distributions $\pi$ and $\mu$ define a probability measure over the space of interaction sequences.

In \cite{Goertzel2010c} this formal agent model is extended in a few ways, intended to make it better reflect the realities of intelligent computational agents.  First, the notion of a {\it goal} introduced, meaning a function that maps finite sequences $ax{s:t}$ into rewards.  As well as a distribution over environments, we have need for a conditional distribution $\gamma$, so that $\gamma(g,\mu)$ gives the weight of a goal $g$ in the context of a particular environment $\mu$.  .  We assume that goals may be associated with symbols drawn from the alphabet $\mathcal G$.     We also introduce a {\it goal-seeking agent}, which is an agent that receives an additional kind of input besides the perceptions and rewards considered above: it receives goals.  

Another modification is to allow agents to maintain memories (of finite size), and at each time step to carry out internal actions on their memories as well as external actions in the environment.   Of course, this could in principle be accounted for within Legg and Hutter's framework by considering agent memories as part of the environment.  However, this would seem an unnecessarily artificial formal model.  Instead we introduce a set $\mathcal C$ of cognitive actions, and add these into the history of actions, observations and rewards.  

Extending beyond the  model given in  \cite{Goertzel2010c}, we introduce here a fixed set of "cognits" $c_i$ (these are atomic cognitions, in the same way that the $p_i$ in the model are atomic perceptions).  Memory is understood to contain a mix of observations, actions, rewards, goals and cognitions.   This extension is a significant one because we are going to model the interaction between atomic cognitions, and in this way model the actual decision-making, action-choosing actions inside the formal agent.   This is big step beyond making a general formal model of an intelligent agent, toward making a formal model of a particular {\it kind} of intelligent agent.   It seems to us currently that this sort of additional specificity is probably necessary in order to say anything useful about general intelligence under limited computational resources.

The convention we adopt is that: When a cognition is  "activated", it acts -- in principle -- on all the other entities in the memory (though in most cases the result of this action on any particular entity may be null).   The result of the action of cognition $c_i$ on the entity $x$ (which is in memory) may be any of:

\begin{itemize}
\item causing $x$ to get removed from the memory ("forgotten")
\item causing some new cognitive entity $c_j$ to get created in (and then persist in) the memory
\item if $x$ is an action, causing $x$ to get actually executed
\item if $x$ is a cognit, causing $x$ to get activated
\end{itemize}

\noindent The process of a cognit acting on the memory may take time, during which various perceptions and actions may occur.   

This sort of cognitive model may be conceived in algebraic terms; that is, we may consider $c_i * x = c_j$ as a product in a certain algebra.   This kind of model has been discussed in detail in \cite{Goertzel1994}, where it was labeled a "self-generating system" and related to various other systems-theoretic models.   One subtle question is whether one allows multiple copies of the same cognitive entity to exist in the memory.   I.e. when a new $c_j$ is created, what if $c_j$ is already in the memory?  Does nothing happen, or is the "count" of $c_j$ in the memory increased?  In the latter case, the memory becomes a multiset, and the product of cognit interactions becomes a (generally quite high dimensional, usually noncommutative and nonassociative) hypercomplex algebra over the nonnegative integers.

In this extended framework, an interaction/internal-action sequence may be written as

$$
c_1 a_1 o_1 g_1 r_1 c_2 a_2 o_2 g_2 r_2  ...
$$

\noindent with the understanding that any of the items in the series may be null.   The meaning of $c_i$ in the sequence is "cognit $c_i$ is activated."    One could also extend the model to explicitly incorporate concurrency, i.e.

$$
c_{11} ... c_{1k_{c 1}} a_{11} ... a_{1k_{a 1}} o_{11} ... o_{1k_{o 1}} g_{11} ... g_{1k_{g 1}} r_{11} ... r_{1k_{r 1}} c_{21} ... c_{2k_{c 2}} a_{21} ... a_{2k_{a 2}} o_{21} ... o_{2k_{o 2}} g_{21} ... g_{2k_{g 2}} r_{21} ... r_{2k_{r 2}} ...
$$

This {\bf Cognit agent} is the next step up in our hierarchy of agents as shown in Figure \ref{fig:agent-classes}.  The next step will be to make the model yet more concrete, by making a more specific assumption about the nature of the cognits being stored in the memory and activated.  

\section{Hypergraph Agents}

Next we assume that the memory of our cognit-based memory has a more specific structure -- that of a {\it labeled hypergraph}.   This yield a basic model of a {\bf Hypergraph Agent} -- a specialization of the Cognit Agent model.

Recall that a hypergraph is a graph in which links may optionally connect more than two different nodes.   Regarding labels: We will assume the nodes and links in the hypergraph may optionally be labeled with labels that are $\textrm{string}$, or structures of the form $(\textrm{string}, \textrm{vector of ints or floats})$.  Here a string label may be interpreted as a node/link type indicator, and the numbers in the vector will potentially have different semantics based on the type.   

Let us refer to the nodes and links of the memory hypergraph, collectively, as Atoms.   In this case the cognits in the above formal model become either Atoms, or sets of Atoms (subhypergraphs of the overall memory hypergraph).   When a cognit  is activated, one or more of the following things happens, depending on the labels on the Atoms in the cognit:

\begin{enumerate}
\item the cognit produces some new cognit, which is determined based on its label and arity -- and on the other cognits that it directly links to, or is directly linked to, within the hypergraph.  Optionally, this new cognit may be activated.
\item the cognit activates one or more of the other cognits that it directly links to, or is directly linked to 
\begin{enumerate}
\item one important example of this is: the cognit, when it is done acting, may optionally re-activate the cognit that activated it in the first place
\end{enumerate}
\item the cognit is interpreted as a {\it pattern} (more on this below), which is then matched against the entire hypergraph; and the cognits returned from memory as "matches" are then inserted into memory
\item in some cases, other cognits may be removed from memory (based on their linkage to the cognit being activated)
\item nothing, i.e. not all cognits can be activated
\end{enumerate}

Option $2a$ allows execution of "program graphs" embedded in the hypergraph.   A cognit $c_1$ may pass activation to some cognit $c_2$ it is linked to, and then $c_2$ can do some computation and link the results of its computation to $c_1$, and then pass activation back to $c_1$, which can then do something with the results.

There are many ways to turn the above framework into a Turing-complete hypergraph-based program execution and memory framework.   Indeed one can do this using only Option 1 in the above list.   Much of our discussion here will be quite general and apply to any hypergraph-based agent control framework, including those that use only a few of the options listed above.   However, we will pay most attention to the case where the cognits include some with fairly rich semantics.   

The next agent model in our hierarchy is what we call an {\bf Rich  Hypergraph Agent}, meaning an agent with a memory hypergraph and a "rich language" of hypergraph Atom types.  In this model, we assume we have  Atom labels for "variable" and "lambda" and "implication" (labeled with a probability value) and "after" (with a time duration).; as well as for "and", "or" and "not", and a few other programmatic operators.   

Given these constructs, we can use a hypergraph some of whose Atoms are labeled "variable" -- such a hypergraph  may be called an "$h$-pattern."  We can also combine $h$-patterns using boolean operations, to get composite $h$-patterns.  We can replicate probabilistic lambda calculus expressions explicitly in our hypergraph.   And, given an $h$-pattern and another hypergraph $H$, we can ask whether $P$ matches $H$, or whether $P$ matches part of $H$.

To conveniently represent cognitive processes inside the hypergraph, it is convenient to include the following labels as primitives: "create Atom" , "remove Atom", plus a few programmatic operations like arithmetic operations and combinators.   In this case the program implementing a cognitive algorithm can be straightforwardly represented in the system hypergraph itself.   (To avoid complexity, we can assume Atom immutability; i.e. make do only with Atom creation and removal, and carry out Atom modification via removal followed by creation.)

Finally, to get reflection, the state of the hypergraph at each point in time can also be considered as a hypergraph.   Let us assume we have, in the rich language, labels for "time" and "atTime."   We can then express, within the hypergraph itself, propositions of the form "At time 17:00 on 1/1/2017, this link existed" or "At time 12:35 on 1/1/2017, this link existed with this particular label."   We can construct subhypergraphs expressing things like "If at time $T$ an subhypergraph matching $P$ exists, then $s$ seconds after time $T$, a subhypergraph matching $P_1$ exists, with probability $p$."   

\subsubsection{The Rich Hypergraph and OpenCog}

The "rich language" as outlined, is in essence a minimal version of the OpenCog AGI system \footnote{see \url{http://opencog.org} for current information, or \cite{EGI1} \cite{EGI2} for theoretical background}.   OpenCog is based on a large memory hypergraph called the Atomspace, and it contains a number of cognitive processes implemented outside the Atomspace which act on the Atomspace, alongside cognitive processes implemented inside the Atomspace.   It also contains a wide variety of Atom types beyond the ones listed above as part of the rich language.   However, translating the full OpenCog hypergraph and cognitive-process machinery into the rich language would be straightforward if laborious.  

The  main reasons for not implementing OpenCog this way now are computational efficiency and developer convenience.   However, future versions of OpenCog could potentially end up operating via compiling the full OpenCog hypergraph and cognitive-process model into some variation on the rich language as described here.   This would have advantages where self-programming is concerned.

\subsection{Some Useful Hypergraphs}

The hypergraph memory we have been discussing is in effect a whole intelligent system -- save the actual sensors and actuators -- embodied in a hypergraph.   Let us call this hypergraph "the system" under consideration (the intelligent system).   We also will want to pay some attention to a larger hypergraph we may call the  "meta-system", which  is created with the same formalism as the system, but contains a lot more stuff.   The meta-system records a plenitude of actual and hypothetical information about the system.

We can represent states of the system within the formalism of the system itself.  In essence a "state" is a proposition of the form "$h$-pattern $P_1$ is present in the system" or "$h$-pattern $P_1$ matches the system as a whole."   We can also represent probabilistic (or crisp) statements about transitions between system states within the formalism of the system, using lambdas and probabilistic implications.  To be useful, the meta-system will need to contain a significant amount of Atoms referring to states of the system, and probabilistically labeled transitions between these states.

The implications representing transitions between two states, may be additionally linked to Atoms indicating the proximal cause of the transition.   For the purpose of modeling cognitive synergy in a simple way, we are most concerned with the case in which there is a relatively small integer number of cognitive processes, whose action reasonably often cause changes in the system's state.   (We may also assume some  can occur for other reasons besides the activity of cognitive processes, e.g. inputs coming into the system, or simply random changes.)   

So for instance if we have two cognitive processes called Reasoning and Blending, which act on the system, then these processes each correspond to a subgraph of the meta-system hypergraph: the subgraph containing the links indicating the state transitions effected by the process in question, and the nodes joined by these links.   This representation makes sense whether or not the cognitive processes are implemented within the hypergraph, or a external processes acting on the system.   We may call these "CPT graphs", short for "Cognitive Process Transition hypergraphs."

\section{PGMC Agents: Intelligent Agents with Cognition Driven by Probabilistic History Mining}

For understanding cognitive synergy thoroughly, it is useful to dig one level deeper and  model the internals of cognitive processes in a way that is finer-grained and yet still abstract and broadly applicable.

\subsection{Cognitive Processes and Homomorphism}

In principle cognitive processes may be very diverse in their implementation as well as their conceptual logic.   The rich language as outlined above enables implementation of anything that is computable.   In practice, however, it seems that the cognitive processes of interest for human-like cognition may be summarized as sets of {\it hypergraph rewrite rules}, of the sort formalized in \cite{baget2002extensions}.   Roughly, a rule of that sort has an input $h$-pattern and an output $h$-pattern, along with optional auxiliary functions that determine the numerical weights associated with the Atoms in the output $h$-pattern, based on combination of the numerical weights in the input $h$-pattern.   

Rules of this nature may be, but are not required to be, homomorphisms.   One conjecture we make, however, is that for the cognitive processes of interest for human-like cognition, {\it most} of the rules involved (if one ignores the numerical-weights auxiliary functions) are in fact either hypergraph homomorphisms, or inverses of hypergraph homomorphisms.  Recall that a graph (or hypergraph) homomorphism is a composition of elementary homomorphisms, each one of which merges two nodes into a new node, in a way that the new node inherits the connections of its parents.     So the conjecture is

\begin{myconjecture}\label{conj:cog-hom}
Most operations undertaken by cognitive processes take the form either of:
\begin{itemize}
\item Merging two nodes into a new node, which inherits its parents' links
\item Splitting a node into two nodes, so that the children's links taken together compose the (sole) parent's links
\end{itemize}
\noindent (and then doing some weight-updating on the product).   
\end{myconjecture}

\subsection{Operations on Cognitive Process Transition Hypergraphs}

One can place a natural Heyting algebra structure on the space of hypergraphs, using the disjoint union for $\sqcup$, the categorial (direct) product for $\sqcap$, and a special partial order called the cost-order, described in \cite{goertzel_heyting}.   This Heyting algebra structure then allows one to assign probabilities to hypergraphs within a larger set of hypergraphs, e.g. to sub-hypergraphs within a larger hypergraph like the system or meta-system under consideration here.   As reviewed in \cite{goertzel_heyting}, this is an intuitionistic probability distribution lacking a double negation property, but this is not especially problematic.

It is worth concretely exemplifying what these Heyting algebra operators mean in the context of CPT graphs.   Suppose we have two CPT graphs $A$ and $B$, representing the state transitions corresponding to two different cognitive processes.

The meet $A \sqcap B$ is a graph representing transitions between conjuncted states of the system (e.g. "System has $h$-pattern P445 and $h$-pattern P7555", etc.).    If $A$ contains a transition between $P_{445}$ and $P_{33}$, and $B$ contains a transition between $P_{7555}$ and $P_{1234}$; then, $A \sqcap B$ will contain a transition between $P_{445} \& P_{7555}$ and $P_{33} \& P_{1234}$.   Clearly, if $A$ and $B$ are independent processes, then the probability of the meet of the two graphs will be the product of the probabilities of the graphs individually

The join $A \sqcup B$ is a graph representing, side by side, the two state transition graphs -- as if we had a new process $A \textrm{or} B$, and a state of this new process could be either a state of $A$, or a state of $B$.  If $A$ and $B$ are disjoint processes (with no overlapping states), then the probability of the join of the two graphs, is the sum of the probabilities of the graphs individually

The exponent $A^B$ is a graph whose nodes are functions mapping states of $B$ into states of $A$.   So e.g. if $B$ is a perception process and $A$ is an action process, each node in $A^B$ represents a function mapping perception-states into action-states.   Two such functions $F$ and $G$ are linked only if, whenever node $b1$ and node $b2$ are linked in $B$, $F(b1)$ and $G(b2)$ are linked in $G$.   I.e. $F$ and $G$ are linked only if  $(F,G)( link(x,y) ) = link(F(x), G(y) )$,  where by $(F,G)(link(x,y))$ one means the set ${F(x), G(y)}$.

So e.g. two perception-to-action mappings $F$ and $G$ are adjacent in $\textrm{action}^\textrm{perception}$ iff, whenever two perceptions $p_1$ and $p_2$ are adjacent, the action $a1=F(p_1)$ is adjacent to the action $a2 = G(p_2)$.   For instance, if

\begin{itemize}
\item $F(\textrm{perception } p)$ = the action of carrying out perception $p$
\item $G(\textrm{perception } p)$ = the action done in reaction to seeing perception $p$
\end{itemize}

\noindent and

\begin{itemize}
\item $p_1$ = hearing the cat
\item $p_2$ = looking at the cat
\end{itemize}

\noindent We then need

\begin{itemize}
\item $F(p_1)$ = the act of hearing the cat (cocking one?s ear etc.)
\item $G(p_2)$ = the response to looking at the cat (raising ones eyes and making a startled expression)
\end{itemize}

\noindent to be adjacent in the graph of actions.   If this is generally true for various $(p_1, p_2)$ then $F$ and $G$ are adjacent in $\textrm{action}^\textrm{perception}$.    Note that $\textrm{action}^\textrm{perception}$ is also the implication $\textrm{perception} \rightarrow \textrm{action}$, where $\rightarrow$ is the Heyting algebra implication.

Finally, according to the definition of cost-based order $A < A_1$ if $A$ and $A_1$ are homomorphic, and the shortest path to creating $A_1$ from irreducible source graph, is to first create $A$.   In the context of CPT graphs, for instance, this will hold if $A_1$ is a broader category of cognitive actions than $A$.   If $A$ denotes all facial expression actions, and $A_1$ denotes all physical actions, then we will have $A < A_1$.

\subsection{PGMC: Cognitive Control with Pattern and Probability}

Different cognitive processes may unfold according to quite different dynamics.  However, from a general intelligence standpoint, we believe there is a common control logic that spans multiple cognitive processes -- namely, adaptive control based on historically observed patterns.   This process has been formalized and analyzed in a previous paper by the author \cite{goertzel2016probabilistic}, where it was called PGMC or "Probabilistic Growth and Mining of Combinations"; in this section we port that analysis to the context of the current formal model.   This leads us to the next step in our hierarchy of agents models, a {\bf PGMC Agent}, meaning an agent with a rich hypergraph memory, and homomorphism/history-mining based cognitive processes.

Consider the subgraph of a particular CPT graph that lies within the system at a specific point in time.   The job of the cognitive control process (CCP) corresponding to a particular cognitive process, is to figure out what (if anything) that cognitive process should do next, to extend the current CPT graph.   A cognitive process may have various specialized heuristics for carrying out this estimation, but the general approach we wish to consider here is one based on pattern mining from the system's history.   

In accordance with our high-level formal agents model, we assume that the system has certain goals, which  manifest themselves as a vector of fuzzy distributions over the states of the system.   Representationally, we may assume a label "goal", and then assume that at any given time the system has $n$ specific goals; and that, for each goal, each state may be associated with a number that indicates the degree to which it fulfills that goal.   

It is quite possible that the system's dynamics may lead it to revise its own goals, to create new goals for itself, etc.   However, that is not the process we wish to focus on here.   For the moment we will assume there is a certain set of goals associated with the system; the point, then, is that a CCP's job is to figure out how to use the corresponding cognitive process to transition the system to states that will possess greater degrees of goal achievement.

Toward that end, the CCP may look at $h$-patterns in the subset of system history that is stored within the system itself.  From these $h$-patterns, probabilistic calculations can be done to estimate the odds that a given action on the cognitive process's part, will yield a state manifesting a given amount of progress on goal achievement.   In the case that a cognitive process chooses its actions stochastically, one can use the $h$-patterns inferred from the remembered parts of the system's history to inform a probability distribution over potential actions.   Choosing cognitive actions based on the  distribution implied by these $h$-patterns can be viewed a novel form of probabilistic programming, driven by fitness-based sampling rather than Monte Carlo sampling or optimization queries -- this is the the "Probabilistic Growth and Mining of Combinations" (PGMC), process described and analyzed in \cite{goertzel2016probabilistic}.

Based on inference from $h$-patterns mined from history, a CCP can then create probabilistically weighted links from Atoms representing  $h$-patterns in the system's current state, to Atoms representing $h$-patterns in potential future states.   A CCP can also, optionally, create probabilistically weighted links from Atoms representing potential future state $h$-patterns (or present state $h$-patterns) to goals.   It will often be valuable for these various links to be weighted with confidence values alongside probability values; or (almost) equivalently with interval (imprecise) probability values \cite{PLN}.

\section{Theory of Stuckness}

In a real-world cognitive system, each CCP will have a certain limited amount of resources, which it can either use for its own activity, or transfer to another cognitive process.    In OpenCog, for instance, space and time resources tend to be managed somewhat separately, which would mean that a pair of floats would be a reasonable representation of an amount of resources.  For our current theoretical purposes, however, the details of the resource representation don't matter much.

Let us say that a CCP, at a certain point in time, is "stuck" if it does not see any high-confidence, high-probability transitions associated with its own corresponding cognitive process, from current state $h$-patterns to future state $h$-patterns that have significantly higher goal-achievement values.   If a CCP is stuck, then it may not be worthwhile for the CCP to spend its limited resources taking any action at that point.   Or, in some cases, it may be the best move for that CCP to transfer some of its allocated resources so some other cognitive process.  This leads us straight on to cognitive synergy.   But before we go there, let us pause to get more precise about how "getting stuck" should be interpreted in this context.

\subsubsection{A Formal Definition of Stuckness}

Let $G_A$ denote the CPT graph corresponding to cognitive process A.   This is a subgraph of the overall cognitive process transition graph of the system, and it may be considered as a category unto itself, with object being the subgraphs, and a Heyting algebra structure.

Given a particular situation $S$ ("possible world") involving the system's cognition, and a time interval $I$, let e.g. $G_A^{S,I}$ denote the CPT graph of $A$ during time interval $I$, insofar as it exists explicitly in the system (not just in the metasystem).

Where $P$ is a $h$-pattern in the system, and $(S,I)$ is a situation/time-interval pair, let $P(S,I)$ denote the degree to which the system displays $h$-pattern $P$  in situation $S$ during time-interval $I$.    Let $g(S,I)$ denote the average degree of goal-achievement of the system in situation $S$ at time during time interval $I$.   Then if we identify a set $\mathcal{I}$ of time-intervals of interest, we can calculate

$$
g(P) = \frac { \sum_{(S,I), I \in \mathcal{I} } g(S,I) P(S,I) } {  \sum_{(S,I), I \in \mathcal{I}} P(S,I)  }
$$

\noindent to be the degree to which $P$ implies goal-achievement, in general (relative to $\mathcal{I}$; but if this set of intervals is chosen reasonably, this dependency should not be sensitive). 

On the other hand, it is more interesting to look at the degree to which $P$ implies goal-achievement across the possible futures of the system as relevant in a particular situation at a particular point in time.   Suppose the system is currently in situation $S$, during time interval $I_S$.   Then $\mathcal{I}$ may be defined, for instance, as a set of time intervals in the near future after $I_S$.   One can then look at

$$
g_{S,I_S, \mathcal{I}}(P) = \frac { \sum_{(S',I), I \in \mathcal{I} } g(S',I) P(S',I) Prob( (S', I) | (S, t)) } {  \sum_{(S',I), I \in \mathcal{I}} P(S',I) Prob( (S', I) | (S, t)) }
$$

\noindent which  measures the degree to which $P$ implies goal-achievement in situations that may occur in the near future after being in situation $S$.    The confidence of this value may be assessed as

$$
c_{S,I_S,\mathcal{I}}(P) = f( \sum_{(S',I), I \in \mathcal{I}} P(S',I) Prob( (S', I) | (S, t)) )
$$

\noindent where $f$ is a monotone increasing function with range $[0,1]$.   This confidence value is a measure of the amount of evidence on which the estimate $g_{S',I_S}(P)$ is based, scaled into $[0,1]$.   

Finally, we may define $e_{C,I_R, S,I_S}(P,I,I_P)$ as the probability estimate that the CCP corresponding to cognitive process $C$ holds for the proposition that: In situation $S$ during time interval $I_S$, if allocated a resource amount in interval $I_R$ for making the choice, $C$ will make a choice leading to a situation in which $P(S,I) \in I_P$ during interval $I$ (assuming $I$ is after $I_S$).   A confidence value  $c_{C,I_R, S,I_S}(P,I,I_P)$ may be defined similarly to $c_{S',t}(P)$ above.   

Given a set $\mathcal{I}$ of time intervals, one can define $e_{C,I_R, S,\mathcal{I}}(P,I,I_P)$  and $c_{C,I_R, S,\mathcal{I}}(P,I,I_P)$ via averaging over the intervals in $\mathcal{I}$.

The confidence with which $C$ knows how to move forward toward the system's goals in situation $S$ at time $t$ may then be summarized as

$$
\textrm{conf}_{C,S,I_S,\mathcal{I}} = max_P \left( g_{S',I_S, \mathcal{I}}(P) c_{S',I_S,\mathcal{I}}(P) e_{C,I_R, S,\mathcal{I}}(P,I,I_P) c_{C,I_R, S,\mathcal{I}}(P,I,I_P)  \right)
$$

\noindent with

$$
\textrm{stuck}_{C,S,I_S,\mathcal{I}}  =  1 - \textrm{conf}_{C,S,I_S,\mathcal{I}} 
$$

\section{Cognitive Synergy: A Formal Exploration}

What we need for "cognitive synergy" between $A$ and $B$ to exist, is for it to be the case that: For many situations $S$ and times $t$, exactly one of $A$ and $B$ is stuck.   

In the metasystem, records of cases where one or both of $A$ or $B$ were stuck, will be recorded as hypergraph patterns.  The set of $(S,t)$ pairs in the metasystem where exactly one of $A$ and $B$ was stuck to a degree of stuck-ness in interval $I_d=(L,U)$, has a certain probability in the set of all $(S,t)$ pairs in the metasystem.   Let us call this set $\textrm{stuck}_{A,B,I_d}$.  

The set $G^{\textrm{stuck}}_{A,B,I_d}$ of CPT graphs $G_A^{S,t}$ , $G_B^{S,t}$  corresponding to the $(S,t)$ pairs in $\textrm{stuck}_{A,B,I_d}$ can also be isolated in the metasystem, and has a certain probability considered as a subgraph of the metasystem (which can be calculated according to the intuitionistic graph probability distribution).   An overall index of cognitive synergy between $A$ and $B$ can then be calculated as follows.   

Let $\mathcal{P}$ be a partition of $[0,1]$ (most naturally taken equispaced).   Then,

$$
\textrm{cog-syn}_{A,B,\mathcal{P}} = \frac { \sum_{I_d \in \mathcal{P} } w_{I_d} Prob( G^{\textrm{stuck}}_{A,B,I_d} ) } {\sum_{I \in \mathcal{P}} w_{I_d} }
$$

\noindent is a quantitative measure of the amount of cognitive synergy between $A$ and $B$.

Extension of the above definition to more than two cognitive processes is straightforward.   Given $N$ cognitive processes, we can look at pairwise synergies between them, and also at triple-wise synergies, etc.    To define triplewise synergies, we can look at  $\textrm{stuck}_{A,B,C,I_d}$, defined as the set of $(S,I)$ where all but one of the three cognitive processes $A$, $B$ and $C$ is stuck to a degree in $I_d$.   Triplewise synergies correspond to cases where the system would be stuck if it had only two of the three cognitive processes, much more often than it's stuck given that it has all three of them.

\subsection{Cognitive Synergy and Homomorphisms }

The existence of cognitive synergy between two cognitive processes will depend sensitively on how these cognitive processes actually work.   However, there are likely some general principles at play here.  For instance we suggest

\begin{myconjecture} \label{conj:homo}
In a PGMC agent operating within feasible resource constraints: If two cognitive processes $A$ and $B$ have a high degree of cognitive synergy between them, then there will tend to be a lot of low-cost homomorphisms between subgraphs of $G_A^{S,t}$ and $G_B^{S,t}$, but {\bf not} nearly so many low-cost isomorphisms.
\end{myconjecture}

The intuition here is that, if the two CPT graphs are too close to isomorphic, then they are unlikely to offer many advantages compared to each other.  They will probably succeed and fail in the same situations.  On the other hand, if the two CPT graphs don't have some resemblance to each other, then often when one cognitive process (say, $A$) gets stuck, the other one (say, $B$) won't be able to use the information produced by $A$ during its work so far, and thus won't be able to proceed efficiently.   Productive synergy happens when one has two processes, each of which can transform the other one's intermediate results, at somewhat low cost, into its own internal language -- but where the internal languages of the two processes are not identical.

Our intuition is that a variety of interesting rigorous theorems likely exist in the vicinity of this informal conjecture.   However, much more investigation is required.

Along these lines, recall Conjecture \ref{conj:cog-hom} above that most cognitive processes useful for human-like cognition, are implemented in terms of rules that are mostly homomorphisms or inverse homomorphisms.   To the extent this is the case, it fits together very naturally with Conjecture \ref{conj:homo}.   

Suppose $G_A^{S,t}$ and $G_B^{S,t}$ each consist largely of records of enacting a series of hypergraph homomorphisms (followed by weight updates), as Conjecture \ref{conj:homo} posits.    Then one way Conjecture \ref{conj2} would happen would be if the homomorphisms in $G_A^{S,t}$ mapped homomorphically into the homomorphisms in $G_B^{S,t}$.   That is, if we viewed $G_A^{S,t}$ and $G_B^{S,t}$ as their own categories, the homomorphisms posited in Conjecture \ref{conj:nat_trans}  would take the form of functors between these two categories.   

\subsection{Cognitive Synergy and Natural Transformations}

Further interesting twists emerge if one views the cognitive process $A$ as associated with a functor $F_A$ that  maps $G^S$ into $G_A^S \subseteq G^S $, which has the property that it maps $G^{S,t}$ into $G_A^{S,t} \subseteq G^{S,t}$ as well.   The functor $F_A$ maps a state transition subgraph of $S$, into a state transition subgraph involving only transitions effected by cognitive process $A$.   So for instance, if $X$ represents a sequence of cognitive operations and conclusions that have transformed the state of the system, then $F_A(X)$ represents the closest match to $X$ in which all the cognitive operations involved are done by cognitive process $A$.   The cost of $F_A(X)$ may be much higher than the cost of $X$, e.g. if $X$ involves vision processing and $A$ is logical inference, then in $F(X)$ all the transitions involved in vision processing need to be effected by logical operations, which is going to be much  more expensive than doing them in other ways.   

A natural transformation $\eta^{A,B}$ from $F_A$ to $F_B$ associates to every object $X$ in $G^S$ (i.e., to every subgraph of the transition graph $G^S$ of the system $S$) a morphism $\eta^{A,B}_X:F_A(X) \rightarrow F_B(X)$ in $G^S$ so that: for every morphism $f: X \rightarrow Y$ in $G^S$ (i.e. every homomorphic transformation from state transition subgraph $X$ to state transition subgraph $Y$) we have $\eta^{A,B}_Y \circ F_A(f) = F_B(f) \circ \eta^{A,B}_X$.  

This leads us on to our final theoretical conjecture:

\begin{myconjecture} \label{conj:nat_trans}
In a PGMC agent operating within feasible resource constraints, suppose one has two cognitive processes $A$ and $B$, which display significant cognitive synergy, as defined above.   Then,

\begin{enumerate}
\item there is likely to be a natural transformation $\eta^{A, B}$ between the functor $F_A$ and the functor $F_B$ -- and also a natural transformation $\eta^{B, A}$ going in the opposite direction
\item the two different routes from the upper left to the bottom right of the commutation diagram corresponding to $\eta^{A, B}$,

 \begin{equation} \label{eqn:nat-trans1}
  \xymatrix@C+2em@R+2em{
   F_A(X) \ar[r]^{F_A(f)} \ar[d]_{\eta^{A,B}_X} & F_A(Y) \ar[d]^{\eta^{A,B}_Y} \\
   F_B(X) \ar[r]_{F_B(f)} & F_B(Y)
  }
 \end{equation}

will often have quite different total costs
\item Referring to the above commutation diagram and the corresponding diagram for $\eta^{B,A}$,

 \begin{equation} \label{eqn:nat-trans1}
  \xymatrix@C+2em@R+2em{
   F_B(X) \ar[r]^{F_B(f)} \ar[d]_{\eta^{B,A}_X} & F_B(Y) \ar[d]^{\eta^{B,A}_Y} \\
   F_A(X) \ar[r]_{F_A(f)} & F_A(Y)
  }
 \end{equation}

-- often it will involve significantly less total cost to 
\begin{itemize}
\item travel from $F_A(X)$ to $F_B(Y)$ via the left-bottom path in Equation \ref{eqn:nat-trans1}, and then from $F_B(Y)$ to $F_A(Y)$ via the right side of Equation \ref{eqn:nat-trans1}; than to
\item  travel from $F_A(X)$ to $F_A(Y)$  directly via the top of Equation \ref{eqn:nat-trans1}
\end{itemize}

That is, often it will be the case that

\begin{equation} \label{eqn:cost}
\begin{array}{l@{}l}
\textrm{cost}(F_A(X) \xrightarrow{\eta^{A,B}_X}  F_B(X))  \\ 
+ \\
\textrm{cost}(F_B(X) \xrightarrow{F_B(f)} F_B(Y)) \\ 
+ \\
\textrm{cost}(F_B(Y) \xrightarrow{\eta^{B,A}_Y}  F_A(Y)  \\
\ll \\
\textrm{cost}( F_A(X) \xrightarrow{F_A(f)} F_A(Y) )
\end{array}
\end{equation}

\end{enumerate}
\end{myconjecture}

Inequality \ref{eqn:cost} basically says that, given the cost weightings of the arrows, it may sometimes be significantly more efficient to get from $F_A(X)$ to $F_A(Y)$ via an indirect route involving cognitive process $B$, than to go directly from $F_A(X)$ to $F_A(Y)$  using only cognitive process $A$.   This is a fairly direct expression of the cognitive synergy between $A$ and $B$ in terms of commutation diagrams.

To make this a little more concrete, suppose $X$ is a transition graph including the new conclusion that Bob is nice, and $Y$ is a transition graph including additionally the even newer conclusion that Bob is helpful.   Then $f$ represents a homomorphism mapping $X$ into $Y$, via -- in one way or another -- adding to the system's memory the conclusion that Bob is helpful.   Suppose $A$ is a cognitive process called "inference" and $B$ is one called "evolutionary learning."   Then e.g. $F_A(X)$ refers to a version of $X$ in which all conclusions are drawn by inference, and $F_B(Y)$ refers to a version of $Y$ in which all conclusions are drawn by evolutionary learning.   The commutation diagram for $\eta^{A,B} =\eta^{ \textrm{inference}, \textrm{evolution} }$, then looks like

 \begin{equation} \label{eqn:nat-trans3}
  \xymatrix@C+8em@R+8em{
   F_\textrm{inference}(\textrm{BobNice}) \ar[r]^{F_\textrm{inference}(f_{\textrm{nice} \rightarrow \textrm{helpful}})} \ar[d]_{\eta^{\textrm{inference},\textrm{evolution} }_\textrm{BobNice}} & F_\textrm{inference}(\textrm{BobHelpful}) \ar[d]^{\eta^{\textrm{inference},\textrm{evolution}}_\textrm{BobHelpful}} \\
   F_\textrm{evolution}(\textrm{BobNice}) \ar[r]_{F_\textrm{evolution}(f_{\textrm{nice} \rightarrow \textrm{helpful}})} & F_\textrm{evolution}(\textrm{BobHelpful})
  }
 \end{equation}

\noindent and the commutation diagram for $\eta^{ \textrm{evolution}, \textrm{inference} }$ looks like

 \begin{equation} \label{eqn:nat-trans4}
  \xymatrix@C+8em@R+8em{
   F_\textrm{evolution}(\textrm{BobNice}) \ar[r]^{F_\textrm{evolution}(f_{\textrm{nice} \rightarrow \textrm{helpful}})} \ar[d]_{\eta^{\textrm{evolution},\textrm{inference} }_\textrm{BobNice}} & F_\textrm{evolution}(\textrm{BobHelpful}) \ar[d]^{\eta^{\textrm{evolution},\textrm{inference}}_\textrm{BobHelpful}} \\
   F_\textrm{inference}(\textrm{BobNice}) \ar[r]_{F_\textrm{inference}(f_{\textrm{nice} \rightarrow \textrm{helpful}})} & F_\textrm{inference}(\textrm{BobHelpful})
  }
 \end{equation}
 
 The conjecture states that, for cognitive synergy to occur, the cost of getting from $F_\textrm{inference}(\textrm{BobNice})$  to $F_\textrm{inference}(\textrm{BobHelpful}) $ directly via the top arrow of Equation \ref{eqn:nat-trans3} would be larger than the cost of getting there via the left and then bottom of Equation \ref{eqn:nat-trans3} followed by the right of Equation \ref{eqn:nat-trans4}.   That is to get from "Bob is nice" to "Bob is helpful", where both are represented in inferential terms, it may still be lower-cost to map "Bob is nice" into evolutionary-programming terms, then use evolutionary programming to get to the evolutionary-programming version of "Bob is helpful", and then map the answer back into inferential terms.

\section{Some Core Synergies of Cognitive Systems: Consciousness, Selves and Others} \label{sec:self} 

The paradigm case of cognitive synergy is where the cognitive processes $A$ and $B$ involved are learning, reasoning or pattern recognition algorithms.   However, it is also interesting and important to consider cases where the cognitive processes involved correspond to different scales of processing, or different types of subsystem of the same cognitive system.  For instance, one can think about: 

\begin{itemize}
\item $A$ = long-term memory (LTM), $B$ = working memory (WM)
\item $A$ = whole-system structures and dynamics, $B$ = the system's self-model
\item $A$ and $B$ are different ''sub-selves'' of the same cognitive system
\item $A$ is the system's self-model, and $B$ is the system's model of another cognitive system (another person, another robot, etc.)
\end{itemize}

Conjecturally and intuitively, it is natural to hypothesize that

\begin{itemize}
\item Homomorphisms between LTM and WM are what ensure that ideas can be moved back and forth from one sort of memory to another, with a loss of detail but not a total loss of essential structure.   
\item Homomorphisms between the whole system's structures and dynamics (as represented in its overall state transition graph) and the structures and dynamics in its self-model, are what make the self-model structurally reflective of the whole system, enabling cognitive dynamics on the self-model to be mapped meaningfully (i.e. morphically) into cognitive dynamics in the whole system, and vice versa
\item Homomorphisms between the whole system in the view of one subself, and the whole system in the view of an other subself, are what enable two different subselves to operate somewhat harmoniously together, controlling the same overall system and utilizing the knowledge gained by one another
\item Homomorphisms between the system's self-model and its model of another cognitive system, enable both  theory-of-mind  type  modeling of others, and learning about oneself by analogy to others (critical for early childhood learning)
\end{itemize}

Cognitive synergy in the form of natural transformations between LTM and WM means that when unconscious LTM cognitive processing gets stuck, it can push relevant knowledge to WM and sometimes the solution will pop up there.  Correspondingly, when WM gets stuck, it can throw the problem to the unconscious LTM processing, and hope the answer is found there, later to bubble up into WM again (the throwing down being according to a homomorphic mapping, and the bubbling up being according to another homomorphic mapping).   As WM is closely allied with what is colloquially referred to as "consciousness" \cite{goertzel2014haracterizing} -- meaning the reflective, deliberative consciousness that we experience when we reason or reflect on something in our "mind's eye" -- this particular synergy appears key to human conscious experience.  As we move thoughts, ideas and feelings back and forth between our focus of attention and the remainder of our mind and memory, we are experiencing this synergy intensively on an everyday basis -- or so the present hypothesis suggests; i.e. that

\begin{itemize}
\item When we pull a memory into attention, or push something out of attention into the "unconscious", we are enacting homomorphisms on our mind's state transition graph  
\item When the unconscious solves a problem that the focus of attention pushed into it, and then the answer comes back into the attentional focus and gets deliberatively reasoned on more, this is the action of the natural transformation between unconscious and conscious cognitive processes -- it's a case where the cost of going the long way around the commutation diagram from conscious to unconscious and back, was lower than the cost of going directly from conscious premise to conscious conclusion.
\end{itemize} 

Cognitive synergy in the form of natural transformations between system and self mean that when the system as a whole cannot figure out how to do something, it will map this thing into the self-model (via a many-to-one homomorphism, generally, as the capacity of the self-model is much smaller), and see if cognitive processes acting therein can solve the problem.  Similarly, if thinking in terms of the self-model doesn't resolve a solution to the problem, then sometimes "just doing it" is the right approach -- which  means mapping the problem the self-model's associated cognitive processes are trying to solve back to the whole system, and letting the whole system try its mapped version of the problem by any means it can find.

Cognitive synergy in the form of natural transformations between subselves means that when one subself gets stuck, it may  map the problem into the cognitive vernacular of another subself and see what the latter can do.   For instance if one subself, which is very aggressive and pushy, gets stuck in a personal relationship issue, it may map this issue into the world-view of another more agreeable and empathic and submissive subself, and see if the latter can find a solution to the problem.   Many people navigate complex social situations via this sort of ongoing switching back and forth between subselves that are well adapted to different sorts of situations \cite{Rowan1990}.

Cognitive synergy in the form of natural transformations between self-model and other-model means that when one get stuck in a self-decision, one can implicitly ask "what would I do if I were this other mind?" ... "what would this other mind do in this situation?"   It also means that, when one can't figure out what another  mind is going to do via other routes, one can map the other mind's situation back into one's self-model, and ask "what would I do in their situation?" ... "what would it be like to be that other mind in this situation?"

In all these cases, we can see the possibility of much the same sort of process as we conjecture to exist between two cognitive processes like evolutionary learning and logical inference.   We have different structures (memory subsystems, models of various internal or external systems, systematic complexes of knowledge and behavior, etc.) associated with different habitual sets of cognitive processes.   Each of these habitual sets of processes may get stuck sometimes, and may need to call out to others for help in getting unstuck.   This sort of request for help is going to be most feasible if the problem can be mapped into the cognitive world of the helper in a way that preserves its essential structure, even if not all its details; and if the answer the helper finds is then mapped back in a similarly structure-preserving way.

Real-world cognitive systems appear to consist of multiple subsystems that are each  more effective at solving certain classes of problems -- subsystems like particular learning and reasoning processes, models of self and other, memory systems of differing capacity, etc.   A key aspect of effective cognition is the ability for these various subsystems to ask each other for help in very granular ways, so that the helper can understand something of the intermediate state of partial-solution that the requestor has found itself in.   This sort of "cognitive synergy" seems to be reflected, in an abstract sense, in certain "algebraic" or category-theoretic symmetries such as we have highlighted here.   

To achieve this abstract modeling of cognitive-process interdependencies in terms of formal symmetries, we have modeled cognitive systems as hypergraphs and made various additional assumptions; however, we suspect that many of these assumptions are not actually necessary for the main conjectures we have proposed to be true in some form.   Our aim has been, not to propose the most general possible model for exploring these ideas, but rather to outline a relatively simple and general model that enables some of the core underlying symmetries to be articulated in a reasonably elegant way.

\section{Cognitive Synergy in the PrimeAGI Design}

The PrimeAGI cognitive architecture \cite{EGI1}\cite{EGI2}, implemented within the OpenCog software platform, works within the "PGMC-driven rich hypergraph memory  model" agent framework outlined above, extending it via introducing a specific set of cognitive processes.   These cognitive processes act on the hypergraph, mapping the nodes and links they find into new nodes and links, or changing the weights of existing nodes and links.   

The particulars of PrimeAGI have been reviewed elsewhere and we will not attempt to give a good summary here, but we will note some of the key cognitive processes.

Let us begin with the key learning and reasoning algorithms:

\begin{itemize}
\item {\bf PLN}: a forward and backward chaining based probabilistic logic engine, based on the Probabilistic Logic Networks formalism
\item  {\bf MOSES}: an evolutionary program learning framework, incorporating rule-based program normalization, probabilistic modeling and other advanced features
\item  {\bf ECAN}: nonlinear-dynamics-based "economic attention allocation", based on spreading of ShortTermImportance and LongTermImportance values and Hebbian learning
\item  {\bf Pattern Mining}: information-theory based greedy hypergraph pattern mining
\item  {\bf Clustering} and  {\bf Concept Blending}: heuristics for forming new ConceptNodes from existing ones
\end{itemize}

The implementation of PrimeAGI in OpenCog is complex, and each of these cognitive processes is implemented in its own way, for a mix of fundamental and historical reasons.   At present some aspects of these cognitive processes are represented within the hypergraph as nodes and links, and other aspects are represented as external software processes; however, there is a design intention to gradually represent all the core cognitive processes of the system within the hypergraph itself.

In \cite{goertzel2016pgmc} the core logic of each of these cognitive processes has been expressed in terms of the PGMC (Probabilistic Growth and Mining of Combinations) framework outlined above.   The control of these cognitive processes does not, at the present time, follow the PGMC logic in any systematic way; however, the PGMC-ization of OpenCog's cognitive processes is planned for 2017-18, and is currently underway.

Detailed explication of cognitive synergy in PrimeAGI in terms of the formalization of cognitive synergy outlined here would be a significant undertaking and we will pursue this in a future paper.   However, the basic concepts are not difficult to outline.

Underlying the manifestation of cognitive synergy in PrimeAGI (and indeed any rich hypergraph based AI system involving logic as well as program execution) is the Curry-Howard correspondence.  In the PrimeAGI context, this gives an isomorphic mapping from the program-execution transition graph created via executing hypergraph nodes containing executable operations ({\it ExecutionOutputLink} in OpenCog syntax), and the transition graph created by logical inferences in PLN logic.   That is, it explains how sub-hypergraphs corresponding to sets of coordinated executable operations can be mapped into sub-hypergraphs corresponding to logical derivations, and vice versa.   

Now let us see some of the places where the potential for cognitive synergy exists:

\begin{itemize}
\item Pattern mining works by growing $h$-pattern hypergraphs into larger ones.   
\item PLN inference can also be used this way, in that one can feed it an $h$-pattern as a premise; but it then searches the space of possible extensions of its premise, using forward and/or backward chaining, and estimates probabilities associated with these extensions, in a different way than the Pattern Miner.   
\item MOSES expands $h$-patterns in yet a different way, in the sense that each of its "demes" (evolutionary subpopulations) begins with a program tree (which can be expressed as an $h$-pattern) as its seed, and then expands this program tree, step by step.   
\end{itemize}

\noindent So: If any of these three processes gets stuck in expanding a certain $h$-pattern usefully, it may  meaningfully ask one or both of the others to help out and try its own heuristics for expansion.   Pattern mining has the power of brute force, MOSES has the creativity of evolution, and PLN has the ingenuity of probabilistic logic; in any given case, one or the other method may prove more capable than the other of finding the interesting extensions of the $h$-pattern in question.

Clustering and concept blending serve to create new concepts, which are then rated as to their quality.   Either may get stuck, in the sense that, when directed to form new concepts in some particular context, they  may persistently fail to form new concepts with decent quality rating.  In this case they  may help each other out.   Clustering  may form new concepts that are then used as properties in the blending process.   Blending may produce new concepts to be clustered.  Conceptually, it seems clear that there is significant synergy between clustering and blending; though this has yet to be studied empirically in a systematic way.

Pattern mining, PLN and MOSES act largely on nodes that are already there in the hypergraph; however, if they get stuck in a certain context, it may be to their benefit to invoke clustering and/or concept blending to form new concepts, new nodes that will then appear in the $h$-patterns that they use in their cognitive processing.  On the other hand, if clustering or blending is performing poorly in a certain context, they may do better to ask PLN or MOSES to form some new links connecting to the nodes they are acting on.  These new links will then be considered by clustering and blending operations, potentially leading them to better results.

Additionally,  in PrimeAGI, the choices of all the cognitive processes are guided by ECAN, in the sense that they choose Atoms to act on, with attention to the ShortTermImportance values of the Atoms, which are adjusted by ECAN's spreading activation and associated processes.   These other cognitive processes also stimulate Atoms that they utilize and find important, increasing the ShortTermImportance of these Atoms incrementally.   The combination of ECAN-based factors and cognitive-process-internal factors in the choice of which Atoms to consider, intrinsically constitutes a form of cognitive synergy.  To wit:

\begin{itemize}
\item When the internal factors within a cognitive process would not have enough information to guide the cognitive process and would lead it to get "stuck", then ECAN (e.g. doing activation spreading and HebbianLink formation based on the recent activity of the cognitive process in question) may provide guidance and help it out.  
\item On the other hand, if a certain set $S$ of Atoms is important, ECAN itself can do only a limited job of figuring out what other Atoms are also going to be important as a consequence.   Having other  cognitive processes act on $S$ will produce new information that will allow ECAN to do its job better, via spreading along the new links that these other cognitive processes create from the Atoms in $S$ to other Atoms.
\end{itemize}

Beyond these synergies between learning and reasoning algorithms, one can see potential in PrimeAGI for synergies between various internal structures and models, as discussed in section \ref{sec:self} above.   

The working memory in OpenCog is associated with a structure called the AttentionalFocus (AF), comprising those Atoms with the highest ShortTermImportance values as determined by ECAN.   Many cognitive processes operate on Atoms in the AF differently than  on the rest of the system's memory hypergraph.   Synergy between AF-based processing and generic memory-based processing is indeed critical for intelligent system functionality.

PrimeAGI explicitly supports two styles of memory representation: local representation (e.g. a ConceptNode labeled "cat"), and more "global" distributed representation (e.g. a network of nodes and links, whose collective pattern of activity represents the system's understanding of what is a "cat").   Most of the work with OpenCog so far has focused on the local representation, but according to the theory underlying the system, both styles of representation will be necessary in order to achieve a high level of general intelligence.   If one looks at manipulation of local representations and manipulation of distributed representations as different cognitive processes, then one can apply the model of cognitive synergy presented here to analyze the situation.   In OpenCog lingo, the process of turning a distributed representation into a localized one is called "map encapsulation"; and the process of turning a localized representation into a distributed one occurs implicitly as a result of integrated PLN and ECAN activity (alongside other cognitive processes).    This suggests that, in the language of Conjecture \ref{conj:nat_trans} one could model

\begin{itemize}
\item $\eta^{\textrm{distributed representation}, \textrm{local representation}}$ = map encapsulation
\item $\eta^{\textrm{local representation}, \textrm{distributed representation}}$ = PLN, ECAN, etc. 
\end{itemize} 

Synergy is also likely to exist between different connectivity patterns in OpenCog's Atomspace, in the context of application to any complex domain.   Part of the theoretical basis for PrimeAGI is the notion of the "dual network" -- hierarchical and heterarchical knowledge networks that are aligned to work effectively together.   Patterns in the system's hypergraphs memory are often naturally arranged in a hierarchy, from more specialized to more general; but also, within each hierarchical level, patterns are also often associated with other patterns with which they share various properties.    The heterarchy helps with the building of the hierarchy, and vice versa.   If we let $A$ denote the cognitive processes of maintaining the hierarchy, and $B$ denote the cognitive processes of maintaining the heterarchy, then the core idea of the "dual network" in PrimeAGI is not just that coupled hierarchical and heterarchical structures exist, but also that the processes $A$ and $B$ of maintaining them interact in a cognitively-synergetic way.

The above arguments regarding cognitive synergy in PrimeAGI have obviously been somewhat "hand-wavy".   To cash them out as precise arguments would require significant explication, as each of these cognitive processes is a complex mathematical entity unto itself -- and the behavior of each of these processes depends, sometimes subtly, on the real-world situations in which it is exercised.   The abstractions presented here are not the end-point of an analysis of cognitive synergy in PrimeAGI, but rather a milestone somewhere near the start of the path.

\section{Next Directions}

We have presented an abstract, relatively formal model of cognitive synergy, in terms of a series of increasingly specific models of intelligent agency.   This work can be extended in two, somewhat opposite directions:

\begin{itemize}
\item Explicating in more detail how cognitive synergy works in the context of specific combinations of PGMC-driven, hypergraph-based cognitive processes -- such as the ones occurring in PrimeAGI and implemented in OpenCog
\item Generalizing and extending the model, e.g. to other sorts of Cognit Agents besides hypergraph-based agents.   There is a growing literature on categorial models of computation, and it seems clear that the core concepts presented here could be elaborated into the context of these more abstract computation models, beyond hypergraphs.   The role of hypergraph homomorphisms in the above discussions could be replaced by more general sorts of morphisms; and probability distributions over other Heyting algebras could be used in place of distributions over hypergraphs; etc.
\end{itemize}

\noindent It might happen that these two research directions converge, in the sense that exploration of more abstract formulations of cognitive synergy might actually end up simplifying the use of cognitive synergy to analyze the interaction of specific cognitive processes, such as pattern mining and evolutionary learning, in specific AI architectures like PrimeAGI.  In any case, in this paper we have just dipped our toe into these rough but fascinating waters; and most of the pure and applied theoretical work in these directions is yet to be done.

\bibliographystyle{alpha}
\bibliography{bbm}

\begin{thebibliography}{GPG13b}

\bibitem[Baa97]{Baars97}
Bernard Baars.
\newblock {\em In the Theater of Consciousness: The Workspace of the Mind}.
\newblock Oxford University Press, 1997.

\bibitem[BF09]{Baars2009}
Bernard Baars and Stan Franklin.
\newblock Consciousness is computational: The lida model of global workspace
  theory.
\newblock {\em International Journal of Machine Consciousness.}, 2009.

\bibitem[BM02]{baget2002extensions}
Jean-Fran{\c{c}}ois Baget and Marie-Laure Mugnier.
\newblock Extensions of simple conceptual graphs: the complexity of rules and
  constraints.
\newblock {\em Journal of Artificial Intelligence Research}, 16:425--465, 2002.

\bibitem[GIGH08]{PLN}
B.~Goertzel, M.~Ikle, I.~Goertzel, and A.~Heljakka.
\newblock {\em Probabilistic Logic Networks}.
\newblock Springer, 2008.

\bibitem[Goe94]{Goertzel1994}
Ben Goertzel.
\newblock {\em Chaotic Logic}.
\newblock Plenum, 1994.

\bibitem[Goe10]{Goertzel2010c}
Ben Goertzel.
\newblock Toward a formal definition of real-world general intelligence.
\newblock In {\em Proceedings of AGI-10}, 2010.

\bibitem[Goe14]{goertzel2014haracterizing}
Ben Goertzel.
\newblock Characterizing human-like consciousness: An integrative approach.
\newblock {\em Procedia Computer Science}, 41:152--157, 2014.

\bibitem[Goe16a]{goertzel2016pgmc}
Ben Goertzel.
\newblock Opencoggy probabilistic programming.
\newblock 2016.
\newblock \url{http://wiki.opencog.org/w/OpenCoggy_Probabilistic_Programming }.

\bibitem[Goe16b]{goertzel2016probabilistic}
Ben Goertzel.
\newblock Probabilistic growth and mining of combinations: A unifying
  meta-algorithm for practical general intelligence.
\newblock In {\em International Conference on Artificial General Intelligence},
  pages 344--353. Springer, 2016.

\bibitem[Goe17]{goertzel_heyting}
Ben Goertzel.
\newblock Cost-based intuitionist probabilities on spaces of graphs,
  hypergraphs and theorems.
\newblock 2017.

\bibitem[GPG13a]{EGI1}
Ben Goertzel, Cassio Pennachin, and Nil Geisweiller.
\newblock {\em Engineering General Intelligence, Part 1: A Path to Advanced AGI
  via Embodied Learning and Cognitive Synergy}.
\newblock Springer: Atlantis Thinking Machines, 2013.

\bibitem[GPG13b]{EGI2}
Ben Goertzel, Cassio Pennachin, and Nil Geisweiller.
\newblock {\em Engineering General Intelligence, Part 2: The CogPrime
  Architecture for Integrative, Embodied AGI}.
\newblock Springer: Atlantis Thinking Machines, 2013.

\bibitem[Hut05]{Hutter2005}
Marcus Hutter.
\newblock {\em Universal {Artificial} {Intelligence}: {Sequential} {Decisions}
  based on {Algorithmic} {Probability}}.
\newblock Springer, 2005.

\bibitem[Leg08]{legg2008machine}
Shane Legg.
\newblock {\em Machine super intelligence}.
\newblock PhD thesis, University of Lugano, 2008.

\bibitem[LH07a]{LeggHutter2007}
Shane Legg and Marcus Hutter.
\newblock A collection of definitions of intelligence.
\newblock In {\em Advances in Artificial General Intelligence}. IOS, 2007.

\bibitem[LH07b]{Legg2007a}
Shane Legg and Marcus Hutter.
\newblock A definition of machine intelligence.
\newblock {\em Minds and Machines}, 17, 2007.

\bibitem[Row90]{Rowan1990}
John Rowan.
\newblock {\em Subpersonalities: The People Inside Us}.
\newblock Routledge Press, 1990.

\end{thebibliography}

\end{document}